**Large Language Model Assisted Transportation Planning through a Differentiable User Equilibrium Layer**

**Eunseung Lee**
Graduate student
Department of Smart City, Chung-Ang University, Seoul, Korea
84 Heukseok-ro, Dongjak-gu, Seoul 156-756, Korea
moohang1@cau.ac.kr

**Gyeongjun Kim**
Graduate student
Department of Smart City, Chung-Ang University, Seoul, Korea
84 Heukseok-ro, Dongjak-gu, Seoul 156-756, Korea
kkjunn7033@cau.ac.kr

**Keemin Sohn*** Corresponding author
Professor
Department of Urban Engineering, Department of Smart City, Chung-Ang University, Seoul, Korea
84 Heukseok-ro, Dongjak-gu, Seoul 156-756, Korea
kmsohn@cau.ac.kr

## Abstract

Transportation planning has traditionally required substantial expertise in network modeling, traffic assignment, and optimization. As a result, strategic analyses such as network design, capacity expansion, and model calibration remain inaccessible to non-expert decision makers. Recent advances in large language models (LLMs) provide an opportunity to bridge this gap by enabling natural-language interaction with complex transportation planning tools. However, existing LLM-based systems primarily serve as interfaces to simulation software and lack the ability to efficiently search, evaluate, and optimize transportation solutions. This study proposes an integrated human-AI transportation planning framework that combines a large language model, a model context protocol (MCP) interface, an open-source transportation planning engine (AequilibraE), and a differentiable user equilibrium (UE) assignment layer. The framework allows non-expert users to execute network analyses and improve solutions for optimization. A differentiable UE layer is developed by directly differentiating the Karush–Kuhn–Tucker (KKT) conditions of the Beckmann traffic assignment problem. The proposed framework is demonstrated through two representative planning tasks: calibration of travel time functions from observed traffic counts and network design under budget constraints. The examples illustrate how differentiable transportation models and LLM agents can jointly democratize transportation planning and substantially reduce barriers between domain expertise and transportation decision making.

## 1. Introduction

Transportation planning and network management require repeated evaluation of traffic assignment outcomes under numerous alternative scenarios. Traditionally, such analyses have been performed by transportation professionals using specialized software packages and optimization techniques. Consequently, the ability to explore transportation policies remains largely restricted to experts possessing extensive modeling capability.

Meanwhile, large language models (LLMs) have recently demonstrated remarkable capabilities in interacting with complex software systems through natural language. Emerging agent architecture enables LLMs to access external computational tools, execute workflows, and perform iterative reasoning. These developments create an opportunity to democratize transportation planning by allowing decision makers without transportation expertise to interact directly with transportation models through conversational interfaces. However, existing LLM-based transportation applications remain largely descriptive. Most systems use LLMs to retrieve information, generate reports, or automate software operations. They rarely incorporate differentiable transportation models capable of supporting optimization-oriented reasoning. Consequently, current frameworks can evaluate candidate solutions but cannot efficiently guide policy makers toward an optimal solution without enumerating alternatives.

This study proposes a new paradigm for transportation planning in which an LLM functions as a planning agent capable of generating, evaluating, and improving transportation policies through direct interaction with transportation models. The proposed framework integrates four components: (1) an LLM serving as a natural-language planner, (2) a model context protocol (MCP) interface connecting the LLM to external tools, (3) the open-source AequilibraE traffic assignment package as an external tool, and (4) a differentiable user equilibrium (UE) assignment layer.

Among these four components, the differentiable assignment layer constitutes the computational core of the framework. Unlike previous approaches that differentiate specific assignment algorithms, the proposed method directly differentiates the optimality conditions of the Beckmann UE problem. The resulting gradients are independent of the traffic assignment solver and can therefore be integrated into arbitrary transportation planning workflows.

The framework enables a non-expert user to specify planning goals using natural language, such as reducing congestion within a budget or calibrating network parameters from observed traffic counts. The LLM translates the request into modeling actions, invokes transportation analyses through MCP, obtains gradients from the differentiable UE layer, and iteratively proposes improved solutions. The differentiable UE layer is the core of the framework for optimization. Without the layer, LLM would enumerate alternatives and heuristically find the best solution by repeating traffic assignments. The present study proposes a robust way to differentiate the UE traffic assignment with respect to planning variables of users.

The contributions of this study are threefold. **First**, we present an integrated LLM-agent architecture for transportation planning. **Second**, we formulate UE assignment as a differentiable and reusable planning component. **Third**, we demonstrate how gradient-enabled transportation models can support non-expert decision making through natural-language interaction. Together, these contributions establish a foundation for AI-assisted transportation planning systems that combine human objectives, transportation theory, and modern AI agents within a unified framework.

The present study is organized as follows. The next section presents literature review regarding the recent development of LLM-assisted transportation planning schemes. The third section presents the methodology for the integrated AI-human transportation planning framework, including methods to set up MCP servers for AequilibraE and to differentiate UE solutions with respect to planning variables. Formulations of differentiable UE layers are also introduced in this section. The fourth section presents experimental set-ups for testing the two representative transportation planning problems. The fifth section presents and discusses the experimental results. The last section draws conclusions and suggests further extensions to the proposed framework.

## 2. Literature review

### 2.1 LLM Applications in Transportation

Qu et al. (2023) emphasized that LLMs may change the role of artificial intelligence in transportation from task-specific prediction models to general-purpose reasoning agents capable of understanding user intent, interpreting heterogeneous information, and supporting decision making. A recent review by Wandelt et al. (2024) further categorized LLM applications in intelligent transportation into perception, prediction, decision support, simulation, and human–machine interaction, highlighting both their promise and their limitations in reliability, grounding, and safety.

One major application area is transportation chatbots and conversational assistants. In this stream, LLMs are used as natural-language interfaces between users and complex transportation systems. Unlike conventional rule-based chatbots, LLM-based assistants can interpret open-ended questions, summarize multimodal information, and guide users through transportation-related decisions. For example, ChatSUMO demonstrates how an LLM can automate the generation and modification of traffic simulation scenarios in SUMO by translating user instructions into executable simulation workflows (Li et al., 2024). More recently, Li et al. (2026) developed ChatSUMO-Agent, that can generate networks and demand, execute simulations, evaluate outputs, and iteratively refine scenarios according to user-defined objectives. The study demonstrates the feasibility of using LLM agents for end-to-end traffic-simulation workflows, but it focuses on simulation construction and refinement rather than differentiable equilibrium-based optimization.

A second stream concerns transportation policy analysis. Because transportation policies are often expressed in long, technical, and unstructured documents, LLMs are well suited for summarization, classification, comparison, and stakeholder-oriented explanation. Generative AI has been argued to improve public engagement in transportation planning by translating technical planning outputs into accessible narratives and by supporting multilingual communication with non-expert stakeholders (Da et al., 2025). However, these studies mainly focus on text interpretation and policy communication rather than direct interaction with transportation models.

A third application area is planning support systems. Traditional transportation planning support systems typically rely on expert-defined workflows, graphical interfaces, and pre-specified model operations. LLMs extend this paradigm by serving as flexible planning agents that can interpret ambiguous user objectives, decompose them into subtasks, and call external tools. Recent work on LLM-based transportation planning has emphasized the potential of generative AI to support scenario generation, demand analysis, accessibility evaluation, and public-facing explanation (Jin and Ma, 2026). Transportation-specific frameworks further connect LLMs with domain-specific analytical tools. TrafficGPT integrates LLMs with traffic foundation models (Zhang et al., 2024), while Open-TI enables an LLM-based agent to invoke external traffic-analysis packages and coordinate analytical workflows (Da et al., 2024). In travel behavior modeling, Liu et al. (2024) examined whether LLMs can reproduce human mode-choice behavior, while Wang et al. (2025) developed agentic LLMs that can mimic human behavior for day-to-day route choice. These studies suggest that LLMs can represent behavioral reasoning and adaptive decision processes, but they do not yet provide a differentiable optimization mechanism for improving network-level planning alternatives.

A fourth related stream is digital twin assistants for transportation systems. Transportation digital twins integrate data, simulation, prediction, and control within a continuously updated virtual representation of physical transportation systems. Recent reviews have identified digital twins as an emerging platform for transport planning, real-time monitoring, and scenario-based policy evaluation (Nag et al., 2025). LLMs can enhance such digital twins by acting as conversational interfaces that explain system states, generate what-if scenarios, and recommend interventions. More generally, LLM-enhanced digital twin research proposes a description–prediction–prescription framework in which LLMs assist in model construction, simulation interpretation, and decision recommendation (Yang et al., 2025). In transportation, this direction is particularly relevant because planning tasks often require iterative interaction between users, network models, simulation outputs, and policy constraints.

Despite these advances, most existing LLM applications in transportation remain descriptive or workflow-oriented. Transportation chatbots mainly answer questions or generate simulation inputs; policy-analysis systems summarize and classify documents; planning-support systems assist with scenario generation; and digital twin assistants explain or monitor system states. Few studies have connected LLM agents with mathematically grounded traffic assignment models and gradient-based optimization. In particular, existing studies have not integrated an LLM planning agent, an open-source traffic assignment engine, and a differentiable user-equilibrium layer into a unified framework for non-expert transportation planning.

### 2.2 Differentiable Traffic Assignment

Historically, there have been attempts to directly differentiate an optimization solution at a lower level with respect to design variables at an upper level. Spall (1992) suggests a method for simultaneous perturbation stochastic approximation (SPSA) to derive the gradient, which could have been applied to solve transportation problems with traffic assignment. Unlike the conventional finite difference method to approximate the gradient, SPSA perturbs all dimensions simultaneously using random perturbation vectors, reducing the required evaluations to just two measurements per iteration regardless of dimension. More recently, decision-focused learning schemes that differentiate black-box functions based on perturb-and-MAP were developed (Mandi et al., 2024). However, such gradient estimates are inherently approximate and may introduce estimation errors into the optimization process.

In parallel with these approximate gradient approaches, transportation researchers have long studied the analytical sensitivity of traffic equilibrium solutions. Early studies derived equilibrium-flow sensitivities with respect to changes in costs and demand and extended them to bilevel transportation problems, providing a theoretical basis for treating user equilibrium as a parameter-dependent solution mapping (Tobin and Friesz, 1988; Patriksson, 2004). Building on these developments, recent advances in differentiable optimization have enabled equilibrium and optimization problems to be incorporated more directly into learning and decision-making pipelines. In this context, recent approaches to differentiable traffic assignment are broadly categorized into unrolled differentiation, fixed-point implicit differentiation, and variational inequality (VI)-based methods.

First, unrolled differentiation of assignment algorithms differentiates through the iterative steps of a traffic assignment solver by applying backpropagation to the computational graph of the algorithm. Representative works include decision-focused learning and differentiable optimization layers (Donti et al. 2017; Amos and Kolter 2017), as well as transport-specific iterative backpropagation methods (Patwary et al. 2023). While these methods are straightforward to implement, they incur computational costs proportional to the number of assignment iterations and produce solver-dependent gradients.

Second, for fixed-point implicit differentiation, the equilibrium solution is represented as a fixed point $x^* = G(x^*, p)$, and gradients are obtained by implicitly differentiating the fixed-point equation. This idea is widely used in implicit layers and equilibrium models (Bai et al. 2019; Gould et al. 2021). In the transportation research field, Liu et al. (2023) applied this technique to differentiate UE assignment layers within a deep learning pipeline. Although this removes the need for unrolling, the resulting gradient depends on the specific choice of the update mapping $G$, which in traffic assignment typically involves shortest-path or all-or-nothing operators that are inherently non-smooth.

Third, a variational inequality (VI) or convex optimization problem for the equilibrium condition is directly differentiated (Larsson & Patriksson, 1994; Friesz et al., 1993). Recently, Li et al. (2020) cast equilibria in games as variational inequality layers and propose both explicit differentiation through unrolled projection iterations and implicit differentiation based on VI sensitivity. These methods operate at the level of the problem definition rather than a specific algorithm, but they are often expressed in operator form and do not explicitly exploit the full structure of the underlying optimization problem. For Beckmann-type UE problems (Beckmann et al., 1956), this projection-based VI differentiation introduces an additional optimization layer through the projection operator.

More recently, Li and Nie (2026) proposed differentiable traffic assignment (DiTA), which constructs a smooth equilibrium-generation process and obtains gradients through automatic differentiation. This differs from directly differentiating the optimality conditions of the original Beckmann formulation.

In contrast to these approaches, the proposed method directly differentiates the reduced KKT system of the original Beckmann problem. Furthermore, no existing study integrates an LLM planning agent, a transportation assignment engine, and a differentiable equilibrium layer into a unified transportation planning framework.

### 2.3 LLM Tool Integration for Transportation Planning

Recent LLM-agent research has enabled language models to interact with external computational tools and coordinate multi-step tasks. Representative studies include ReAct, which combines reasoning and tool execution (Yao et al., 2023), Toolformer, which enables language models to invoke external tools during generation (Schick et al., 2023), and HuggingGPT, which demonstrates orchestration across multiple external models (Shen et al., 2023).

Such tool integration is particularly relevant to transportation planning, where a single planning request may involve problem formulation, traffic assignment, performance evaluation, sensitivity analysis, and iterative modification of design variables. Standardized tool interfaces can support this process by separating language-based reasoning from specialized numerical computation. The Model Context Protocol (MCP), for example, exposes external computational functions to an LLM through predefined input and output schemas (Anthropic, 2024), while numerical computation remains within dedicated analytical software.

Existing transportation-oriented LLM systems have demonstrated the ability to coordinate simulation or analytical workflows, including simulation scenario generation and heterogeneous traffic-analysis procedures (Li et al., 2024; Zhang et al., 2024; Da et al., 2024). However, equilibrium sensitivity has not yet been exposed as a reusable tool that an LLM planning agent can invoke to improve transportation-planning decisions.

Accordingly, an important remaining gap lies at the intersection of LLM-based tool orchestration and differentiable transportation modeling. A standardized tool-integrated framework can bridge this gap by enabling an LLM to coordinate problem formulation and iterative planning decisions while delegating traffic assignment and equilibrium sensitivity analysis to mathematically grounded transportation models.

## 3. Methodology

### 3.1 Framework Overview

The proposed framework enables non-expert users to perform transportation planning tasks through natural language interaction with an LLM. Unlike conventional transportation planning systems that require extensive knowledge of traffic assignment and optimization, the proposed framework allows users to specify planning objectives using natural language while the underlying transportation analyses are executed automatically.

The framework consists of four major components: (i) an LLM planning agent, (ii) a MCP interface, (iii) the AequilibraE traffic assignment engine, and (iv) a differentiable UE layer. Fig. 1 illustrates the overall architecture.

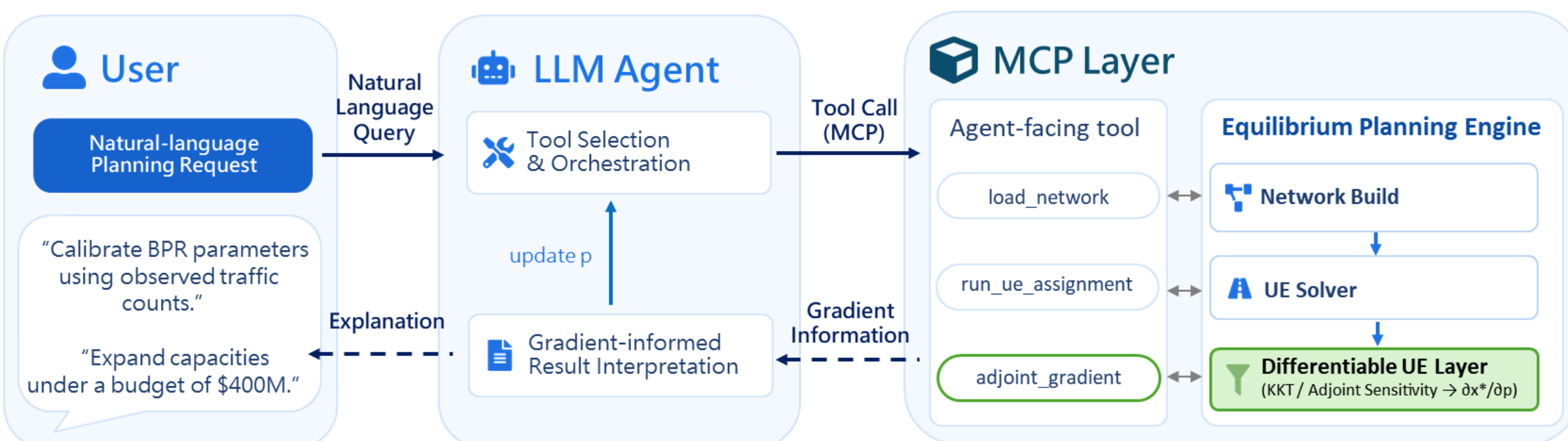

Fig. 1 LLM-assisted transportation planning framework for non-experts

A user first submits a planning request in natural language. The LLM interprets the request and translates it into structured transportation-planning tasks. Through MCP, the LLM invokes transportation analysis functions implemented in AequilibraE that is an open-source transportation planning tool. The resulting network states and performance measures are returned to the LLM. When optimization or calibration is required, gradients obtained from the differentiable UE layer are used to iteratively improve the planning alternatives. Consequently, transportation planning is formulated as an interactive human–AI decision-making process in which domain expertise is partially transferred from transportation specialists to the AI agent.

### 3.2 LLM Transportation Planning Agent

The LLM serves as a planning agent that translates user objectives into transportation optimization problems. A user request is first converted into a structured planning representation $P=\{O,C,B,S\}$, where O denotes the planning objective, C denotes constraints, B denotes available resources, and S denotes the current transportation system state.

Typical examples can include:

- Provide additional capacities to reduce congestion with a limited budget
- Identify the most critical bottlenecks
- Calibrate network parameters using observed traffic counts

The planning agent automatically determines the required transportation analyses and invokes the corresponding MCP tools. Unlike conventional transportation planning support systems that merely execute predefined workflows, the proposed agent iteratively updates planning alternatives using transportation-model feedback. The planning process therefore resembles a closed-loop interaction between natural-language reasoning and transportation-system simulation.

### 3.3 MCP-Based Transportation Planning Interface

The MCP provides the interface between the LLM planning agent and the numerical transportation-analysis modules. Its role is not to solve the transportation problem itself, but to expose verified transportation functions to the LLM through predefined tool interfaces. After the planning agent interprets a natural-language request and constructs the planning representation described in Section 3.2, it selects and invokes the MCP tools required to execute the corresponding analysis.

Table 1 summarizes the MCP tools implemented in this study. The tools are organized into six providers according to their computational roles. The network data provider initializes the transportation problem by loading the network, OD demand, and, when required, observed traffic counts. The UE assignment provider executes verified UE assignment and evaluates the resulting network state or task-specific objective. These operations form the forward transportation-modeling component of the framework.

The differentiable UE provider exposes the gradient of the planning objective with respect to upper design variables, which is developed in Section 3.4. After a UE solution has been obtained, adjoint_gradient uses the converged equilibrium state to construct and solve the reduced-KKT adjoint system. It therefore returns the gradient of the planning objective without requiring a separate UE assignment for each design variable perturbation. This distinction is particularly important for high-dimensional problems, because the entire gradient vector can be obtained from the equilibrium state and one adjoint calculation rather than through repeated perturbation-based assignments.

Two task-specific providers support the applications considered in this study. For BPR parameter calibration, the BPR calibration provider configures the estimation problem and coordinates calibration using the differentiable UE layer. For the continuous network design problem, the network design provider checks budget and bound constraints and projects candidate capacity-expansion vectors onto the feasible design set. These providers encapsulate transportation-specific numerical operations while leaving the LLM responsible for interpreting the user's objective, selecting appropriate tools, and coordinating the overall workflow.

Finally, the reporting provider maintains the computational history and returns the best feasible solution obtained during the planning process. This allows the LLM to retrieve structured information about previous evaluations rather than reconstructing the optimization history from unstructured console output. The resulting interaction follows a closed-loop sequence in which the LLM selects an action, MCP invokes the corresponding transportation tool, the numerical backend returns a structured result, and the LLM uses that result to determine the next action.

An important feature of this architecture is the separation between language-based reasoning and numerical transportation computation. The LLM does not directly calculate UE flows, equilibrium sensitivities, or feasibility projections. These operations are delegated to verified numerical routines, while MCP defines how their inputs and outputs are communicated to the agent. Consequently, the same UE and adjoint routines can be accessed either through MCP or through another interface, such as the command-line interface used by Gradient-Agent in the experiments. MCP therefore does not alter the mathematical solution or the cost of an individual UE or adjoint computation; its potential benefit lies in providing structured, task-specific access to these capabilities and in reducing unnecessary tool interactions during an iterative planning workflow.

Because each provider is exposed through a predefined input-output schema, the numerical backend can in principle be replaced by another transportation package if equivalent tool functions are implemented while preserving the same MCP interface. In the present implementation, AequilibraE provides the verified forward UE assignment functionality, whereas the reduced-KKT differentiation and task-specific optimization utilities are implemented as dedicated modules of the proposed framework.

Table 1. MCP tool interface of the differentiable UE planning framework.

| Provider | MCP tool | Role |
|---|---|---|
| **Network Data Provider** | `load_network, load_counts` | Initializes the network, demand, observed counts. |
| **UE Assignment Provider** | `run_ue_assignment, evaluate` | Solves user equilibrium and returns the objective or count-fit metrics. |
| **Differentiable UE Provider** | `adjoint_gradient` | Computes the reduced-KKT adjoint gradient from the current converged UE state without UE re-evaluation. |
| **Network Design Provider** | `check_design, project_design` | Checks budget feasibility and projects a design onto the feasible set. |
| **BPR Calibration Provider** | `set_calibration_config, run_calibrate, split_links` | Estimates link-level BPR parameters through the differentiable UE layer. |
| **Reporting Provider** | `get_history, get_solution` | Returns the tool-call log and the best-so-far solution. |

### 3.4 Differentiable User Equilibrium Layer

#### 3.4.1 Beckmann UE problem

Let $L$ denote the set of directed links and $x = (x_a)_{a\in L}$ the link-flow vector. Let $p$ denote a generic parameter vector, such as link-performance function parameter, OD demand, link capacity, free-flow travel time, or network design variables. For a given $p$, Eq. (1) denotes the Beckmann problem that represents deterministic UE.

$$x^*(p) = \arg\min_{x\in\mathcal{X}(q)} B(x,p) \text{ where } B(x,p) = \sum_{a\in L} \int_0^{x_a} t_a(u,p)\, du \quad (1)$$

Here, let $\mathcal{X}(q)$ denote the feasible set of aggregate link flows induced by the fixed OD matrix $q$.

#### 3.4.2 Setting-up differentiable KKT system

The Lagrangian of Beckmann UE problem is represented by Eq. (2).

$$\mathcal{L}(x,\lambda,\mu,p) = B(x,p) + \lambda^\top(Ax - b(p)) - \mu^\top x \quad (2)$$

The KKT conditions for the problem are listed in Eq. (3).

$$\begin{aligned} &t(x,p) + A^\top\lambda - \mu = 0,\\ &Ax - b = 0,\\ &x \ge 0, \mu \ge 0,\\ &\mu \odot x = 0. \end{aligned} \quad (3)$$

Here, $A$ denotes a full-row-rank local representation of the feasible aggregate-flow set induced by the fixed OD demand matrix, rather than the ordinary node-link incidence matrix based only on aggregate productions and attractions. The underlying OD-specific conservation conditions are implicitly embedded in this reduced representation. Accordingly, the differentiated KKT system can be expressed in terms of aggregate link flows without explicitly retaining OD-specific flow variables. $b$ represents the demand-dependent right-hand side. $\odot$ denotes the Hadamard product. The last condition represents complementarity, and how to treat this discrete condition is described in this subsection.

Then the KKT system can be compactly written as Eq. (4).

$$\mathcal{F}(x, \lambda, \mu, p) = 0 \tag{4}$$

Here, $\mathcal{F}(x, \lambda, \mu, p) = \begin{bmatrix} t(x,p) + A^{\top}\lambda - \mu \\ Ax - b \\ \mu \odot x \end{bmatrix}$. The inequalities $x \geq 0$ and $\mu \geq 0$ are imposed separately.

**(1) Treatment of complementarity via local active-set identification**

The active-set needs to be defined to treat complementarity in a plausible way. Let $x^*(p)$ be the Beckmann UE solution and define the active positive-flow link set as

$$\mathcal{A}(p) = \{\, a \in L \mid x_a^*(p) > 0 \,\}.$$

The positive-flow set determines the locally active nonnegativity constraints, while the OD-feasibility structure is retained through the local representation of $\mathcal{X}(q)$. Its complement as

$$\mathcal{Z}(p) = L \setminus \mathcal{A}(p).$$

For $a \in \mathcal{A}(p)$ we have $\mu_a^*(p) = 0$, while for $a \in \mathcal{Z}(p)$ we have $x_a^*(p) = 0$.

We assume that the active set is locally invariant around the current parameter value $p$:

$$\exists\, \epsilon > 0 \text{ such that } \mathcal{A}(p + \delta p) = \mathcal{A}(p) \;\; \forall \| \delta p \| < \epsilon.$$

This is a standard regularity condition that holds away from degeneracy or changes in the locally active feasible set. Under strict complementarity,

$$x_a^*(p) > 0 \Rightarrow \mu_a^*(p) = 0, \qquad x_a^*(p) = 0 \Rightarrow \mu_a^*(p) > 0,$$

the solution is locally stable and the active set remains unchanged in a neighborhood of $p$.

Under this assumption, the original KKT conditions [Eq. (3)] can be reduced to a smooth system of equalities on the locally fixed active set.

For $a \in \mathcal{A}(p)$:

$$t_a(x,p) + (A^{\top}\lambda)_a = 0, \qquad x_a > 0, \qquad \mu_a = 0.$$

For $a \in \mathcal{Z}(p)$:

$$x_a = 0, \qquad \mu_a = t_a(x,p) + (A^{\top}\lambda)_a \geq 0.$$

Eliminating $\mu$ yields the reduced system in variables $(x_{\mathcal{A}}, \lambda)$ as shown in Eq. (5)

$$\begin{aligned} &t_{\mathcal{A}}(x,p) + A_{\mathcal{A}}^{\top}\lambda = 0, \\ &Ax = b, \\ &x_{\mathcal{Z}} = 0. \end{aligned} \tag{5}$$

Here, $A_{\mathcal{A}}$ denotes the submatrix of $A$ restricted to links in $\mathcal{A}$. The nonnegativity condition $\mu \geq 0$ is implicitly enforced through the active set identification. Under the assumption that the active set remains locally unchanged, the inequality constraints remain satisfied and do not explicitly appear in the differentiated system. After identifying the positive-flow active set, the corresponding reduced feasibility matrix $A_{\mathcal{A}}$ is obtained by restricting the local representation to the active links.

**(2) Differentiability**

Using the proposed framework, the equilibrium flow is treated as a differentiable mapping $x^*(p)$, defined implicitly by the reduced KKT system of the Beckmann UE problem. Under the local active-set stability assumption, the complementarity conditions can be eliminated and the KKT system reduces to Eq. (6) that is a smooth system in $(x_{\mathcal{A}}, \lambda)$.

$$\tilde{\mathcal{F}}(x^*_{\mathcal{A}}(p), \lambda^*(p), p) = 0, \tag{6}$$

where $\tilde{\mathcal{F}} = \begin{bmatrix} t_{\mathcal{A}}(x^*(p), p) + A_{\mathcal{A}}^{\top}\lambda^*(p) \\ A_{\mathcal{A}} x^*_{\mathcal{A}}(p) - b \end{bmatrix}$.

Under local active-set stability and nonsingularity of the reduced KKT Jacobian, the implicit function theorem establishes a locally differentiable equilibrium mapping. The inactive flows are fixed as $x^*_{\mathcal{Z}}(p) = 0$. The implicit differentiation leads to Eq. (7).

$$\tilde{\mathcal{F}}_{(x_{\mathcal{A}},\lambda)} \frac{d[x_{\mathcal{A}},\lambda]}{dp} + \tilde{\mathcal{F}}_p = 0 \tag{7}$$

The matrix inverse emerges to derive the Jacobian of $(x_{\mathcal{A}}, \lambda)$ with respect to $p$. Eq. (8) denotes the Jacobian provided by the differentiable UE layer.

$$\frac{d[x_{\mathcal{A}},\lambda]}{dp} = -\tilde{\mathcal{F}}^{-1}_{(x_{\mathcal{A}},\lambda)} \tilde{\mathcal{F}}_p \tag{8}$$

Here, $\tilde{\mathcal{F}}_{(x_{\mathcal{A}},\lambda)} = \begin{bmatrix} D_{t,\mathcal{A}} & A_{\mathcal{A}}^{\top} \\ A_{\mathcal{A}} & 0 \end{bmatrix}$ with $D_{t,\mathcal{A}} = \operatorname{diag}\left(\frac{\partial t_a(x_a, p_a)}{\partial x_a}\right)_{a \in \mathcal{A}}$.

This reduced representation avoids explicitly differentiating the complementarity condition and removes the multiplier $\mu$ from the differentiated system.

When the planning objective function is defined as $J(p) = \Phi(x^*(p))$, the gradient is represented by Eq. (9).

$$\nabla_p J = \Phi_p - \Phi_{(x_{\mathcal{A}},\lambda)} \tilde{\mathcal{F}}^{-1}_{(x_{\mathcal{A}},\lambda)} \tilde{\mathcal{F}}_p \tag{9}$$

Here, the adjoint system is defined as Eq. (10), and the gradient is given by Eq. (11).

$$\tilde{\mathcal{F}}^{\top}_{(x_{\mathcal{A}},\lambda)} \psi = \Phi^{\top}_{(x_{\mathcal{A}},\lambda)} \tag{10}$$

$$\nabla_p J = \Phi_p - \psi^{\top} \tilde{\mathcal{F}}_p. \tag{11}$$

This adjoint formulation converts the matrix inversion into solving the adjoint system, which can be implemented in a much simpler way by using matrix decomposition scheme.

The mapping $x^*(p)$ is generally piecewise smooth. The above derivative is valid within a region where the active set is fixed. The mapping is locally differentiable as long as the active feasible set remains unchanged and the reduced KKT Jacobian is nonsingular. However, changes in the active set may introduce nondifferentiable points.

In practice, the active set $\mathcal{A}(p)$ is identified from the converged UE solution $x^*(p)$ by thresholding $\mathcal{A}(p) = \{a \mid x_a^*(p) > \delta\}$, for a small tolerance $\delta > 0$. The reduced system is then constructed and differentiated.

### 3.5 Gradient-Guided Search

Once the differentiable UE layer is available, transportation planning can be formulated as a continuous optimization problem. Let $p$ denote planning variables and let $J(p)$ represent a transportation performance measure such as total system travel time or calibration error.

When a planning problem is given, Eq. (12) denotes the gradient of the objective function represented by the chain rule.

$$\nabla_p J = \left(\frac{d[x_{\mathcal{A}},\lambda]}{dp}\right)^T \nabla_{[x_{\mathcal{A}},\lambda]} J = \Phi_p - \psi^\top \tilde{\mathcal{F}}_p \quad (12)$$

The LLM uses this information to identify promising planning directions and generate improved alternatives. Rather than exploring alternatives through exhaustive search or trial-and-error simulation, the proposed framework allows LLMs to efficiently guide the planning process.

### 3.6 Human–AI Transportation Planning Workflow

The complete planning procedure consists of six steps. The workflow is repeated until the user accepts a planning solution or a stopping criterion is satisfied. This iterative interaction establishes a human–AI collaborative planning process in which transportation expertise is embedded within the computational framework while decision authority remains with the user.

Step 1. The user specifies a planning goal using natural language.

Step 2. The LLM interprets the request and formulates a transportation-planning problem.

Step 3. MCP invokes the required transportation-analysis tools in AequilibraE.

Step 4. UE traffic assignment is performed and network-performance measures are evaluated.

Step 5. The differentiable UE layer computes equilibrium sensitivities and objective gradients.

Step 6. The LLM generates improved planning alternatives and presents them to the user.

## 4. Experimental set-ups

Two transportation-planning tasks are considered to evaluate whether an LLM agent can perform optimization-oriented analyses that traditionally require specialized knowledge of traffic assignment and mathematical programming. The first task (Task1) is the calibration of link performance functions from observed traffic counts. Although traffic counts may be readily available to practitioners, estimating network-wide link performance parameters requires repeated equilibrium assignment and solution of an inverse optimization problem. The proposed framework is intended to allow non-expert users to carry out this task without directly implementing traffic-assignment or optimization algorithms.

The second task (Task2) is a continuous network design problem (CNDP), in which capacity expansions are allocated across candidate links under a capital-budget constraint. Such decisions require not only identifying congested links but also evaluating how local capacity changes redistribute equilibrium flows throughout the network and determining whether their system-wide benefits justify the associated investment costs. The two tasks therefore represent complementary applications of the proposed framework: inverse calibration of an existing network model and forward optimization of network capacity.

### 4.1 Target Network and Agent Configurations

Both tasks are tested on the Sioux Falls network, a standard benchmark in the traffic-assignment literature. The network consists of 24 nodes, 76 directed links, and 528 OD pairs with positive demand. The same basic network is used for both tasks, while task-specific experimental conditions are introduced separately in Sections 4.2 and 4.3. The network provides sufficient congestion and route interaction for changes in model parameters or link capacities to induce meaningful redistribution of UE flows.

Four agent configurations are evaluated, as summarized in Table 2. For comparisons conducted with a given LLM, all four agents receive the same network, demand, and task-specific input data and differ in the transportation-modeling capabilities made available to them. This design allows the effects of providing an agent with UE assignment, differentiable UE layer, and MCP-based tool integration to be examined sequentially. The experiments are subsequently repeated across four different LLMs to examine whether these effects depend on the underlying language-model capability.

Table 2. Experimental design: agent configurations and available capabilities.

| Agent | MCP | Differentiable UE | UE Assignment | Network & OD Data |
|---|---|---|---|---|
| **Proposed-Agent** | ✓ | ✓ | ✓ | ✓ |
| **Gradient-Agent** | – | ✓ | ✓ | ✓ |
| **UE-Agent** | – | – | ✓ | ✓ |
| **Naive-Agent** | – | – | – | ✓ |

Naive-Agent receives the network, OD demand matrix, and task-specific data but has no access to an established transportation-assignment package. General code execution is permitted, so the agent may construct its own traffic-assignment and optimization procedures. This condition represents a general-purpose LLM attempting to solve the transportation-planning task without a verified domain-specific assignment tool.

UE-Agent has access to the verified UE assignment routines of AequilibraE through a standard API but receives no gradient information. The agent must therefore determine its own upper-level search strategy, for example by using numerical perturbations or heuristic search. Comparison between UE-Agent and Naive-Agent assesses the value of providing a verified transportation-assignment engine, whereas comparison between UE-Agent and the gradient-equipped agents evaluates the additional contribution of the differentiable UE assignment layer.

Gradient-Agent has access to both the verified UE assignment engine and the differentiable UE layer developed in Section 3.4. These functions are invoked through a command-line interface (CLI), and the exact gradient of the planning objective is obtained from the reduced-KKT adjoint system. The agent can therefore perform gradient-guided search without constructing an approximate UE sensitivity by itself.

Proposed-Agent is provided with the same verified UE assignment and differentiable UE capabilities as Gradient-Agent, but these functions are exposed to the LLM through the MCP interface described in Section 3.3. Thus, comparison between Proposed-Agent and Gradient-Agent primarily examines whether structured MCP-based tool access affects the way in which an LLM orchestrates repeated transportation-model evaluations, rather than the numerical quality of the underlying UE or gradient computation.

The four agent configurations are evaluated using four different LLMs: GPT-Luna, GPT-Sol, Claude Sonnet, and Claude Opus. Comparisons among agent configurations within the same LLM isolate differences associated with tool access, whereas comparisons across LLMs examine the extent to which agent performance depends on the capability of the underlying language model.

### 4.2 Experimental set-up for BPR parameter calibration (Task 1)

We consider an inverse equilibrium problem in which the parameters of the link performance functions are estimated so that the UE link flows reproduce observed link counts as closely as possible.

#### 4.2.1 Definition of Task 1

Let $\theta$ denote the vector of BPR parameters to be estimated. For each link $a \in L$, the travel time function is written as Eq. (13).

$$t_a(x_a, \theta_a) = t_a^0 \left[1+\alpha_a \left(\frac{x_a}{C_a}\right)^{\beta_a}\right], \text{ where } \theta_a = (\alpha_a, \beta_a). \tag{13}$$

Let $x^{obs}$ denote the observed link-flow vector, and let $H$ be a selection matrix that extracts the observed links. The observed link counts are therefore represented by $Hx^{obs}$. For a given parameter vector $\theta$, the UE flow is defined by the Beckmann formulation. Eq. (14) denotes the parameter estimation problem.

$$\min_{\theta} J(\theta) = \frac{1}{2} \| Hx^*(\theta) - x_H^{obs} \|_2^2 + \frac{\rho}{2} \| \theta - \theta^0 \|_2^2 \tag{14}$$

Here, the BPR parameters are constrained by $\alpha_a \geq 0$ and $\beta_a \geq 1$ for $\forall a \in A$. $x_H^{obs}$ is the vector of observed link counts on the monitored links, $\theta^0$ is a prior or nominal BPR parameter vector, and $\rho \geq 0$ controls the strength of regularization.

Using the reduced KKT representation, the equilibrium flow is defined implicitly by Eqs. (15) and (16).

$$\tilde{\mathcal{F}}(x_{\mathcal{A}}^*(\theta), \lambda^*(\theta), \theta) = 0 \tag{15}$$

$$x_{\mathcal{Z}}^*(\theta) = 0 \tag{16}$$

Eq. (17) denotes the gradient of the objective obtained from the adjoint system.

$$\tilde{\mathcal{F}}_y^{\top} \psi = \Phi_y^{\top} \tag{17}$$

Here, $\Phi(x, \theta) = \frac{1}{2} \| Hx - x_H^{obs} \|_2^2 + \frac{\rho}{2} \| \theta - \theta^0 \|_2^2$.

Thus, $\Phi_x = H^{\top}(Hx^*(\theta) - x_H^{obs})$ and $\Phi_\theta = \rho(\theta - \theta^0)$. Eq. (18) denotes the final gradient.

$$\nabla_\theta J = \Phi_\theta - \psi^\top \tilde{\mathcal{F}}_\theta \qquad (18)$$

For the BPR function, the required derivatives with respect to parameters are computed with ease.

This application demonstrates how the proposed differentiable UE layer can be used for inverse calibration, instead of directly fitting link travel time functions. The parameters are estimated so that the equilibrium flows induced by those parameters match observed network traffic patterns.

#### 4.2.2 Experimental settings for Task 1

(1) Synthetic ground-truth parameters and observed traffic counts

The ground-truth coefficients are independently sampled according to Eq. (19),

$$\alpha_a \sim \mathcal{N}(0.15,\ 0.05^2) \text{ and } \beta_a \sim \mathcal{N}(4.0,\ 0.5^2) \qquad (19)$$

and clipped to $\alpha_a \in [0.05, 0.40]$ and $\beta_a \in [3.0, 5.5]$. The clipping prevents extreme parameter values while retaining substantial heterogeneity across links.

Using these ground-truth parameters, UE assignment is performed with the conjugate Frank–Wolfe algorithm to a relative gap of $3.3 \times 10^{-7}$. The resulting equilibrium flows are treated as synthetic observed traffic counts. No additional measurement noise is introduced, allowing the experiment to isolate differences in the agents' ability to solve the inverse equilibrium problem rather than differences arising from measurement error.

Partial network observation is introduced by selecting 58 links whose equilibrium volume-to-capacity ratios satisfy $V/C \geq 1$ under the ground-truth configuration. This creates a calibration setting in which observations are concentrated on congested links while a substantial portion of the network remains unmonitored. Only the selected link flows are supplied as observations to the agents; the synthetic ground-truth BPR parameters are retained solely for ex-post evaluation.

(2) Estimation settings

Table 3 summarizes the common estimation settings. The nominal coefficients $\alpha$ = 0.15 and $\beta$ = 4.0 are used as the prior values in the regularization term. To ensure that calibration does not begin at the prior, initial parameter values are independently generated around $\alpha$ = 0.30 and $\beta$ = 5.0 and subsequently restricted to the prescribed parameter bounds.

All agents are allowed a maximum of 1,500 outer calibration iterations. Adam is used for the gradient-based calibration procedures, with learning rates of 0.002 for $\alpha$ and 0.01 for $\beta$. The regularization coefficient is set to $\rho$ = 1.5. The parameter ranges are restricted to $\alpha \in$ [0.05, 0.40] and $\beta \in$ [3.0, 5.5] throughout the estimation.

The number of UE assignments is not capped within this common optimization horizon and is instead recorded as an outcome measuring computational effort. This distinction is important because agents without the differentiable UE layer may perform multiple UE flow evaluations to construct a search direction, whereas one adjoint evaluation provides the full parameter gradient for the differentiable UE layer.

Table 3. Experimental settings for BPR parameter calibration.

| Item | Value | Item | Value |
|---|---|---|---|
| **Network** | Sioux Falls<br>(24 nodes, 76 links, 528 positive-demand OD pairs) | **Maximum outer iterations** | 1500 |
| **Monitored links** | 58 (ground-truth V/C ≥ 1.0) | **Optimizer** | Adam |
| **Active links** | 76 | **Learning rates** | $\eta_\alpha = 0.002$, $\eta_\beta = 0.01$ |
| **Prior values** | $\alpha$ = 0.15; $\beta$ = 4.0 | **Regularization $\rho$** | 1.5 |
| **Initial values** | $\alpha \sim \mathcal{N}(0.30, 0.05^2)$, $\beta \sim \mathcal{N}(5.0, 0.5^2)$ | **Clipping range** | $\alpha \in [0.05, 0.4]$, $\beta \in [3.0, 5.5]$ |

### 4.2.3 Evaluation metrics for Task 1

The BPR calibration experiment is evaluated from four perspectives: flow-reproduction accuracy, parameter-recovery accuracy, computational efficiency, and run-to-run consistency. Table 4 summarizes the corresponding metrics. Flow-reproduction accuracy is the primary criterion because the optimization objective is to reproduce observed traffic counts through the equilibrium model. Root-mean-square error (RMSE) and mean absolute percentage error (MAPE) are calculated between the observed and assigned flows on the 58 monitored links. In addition, the GEH statistic is reported as a conventional traffic-model validation measure. The proportion of monitored links satisfying GEH < 5 is reported, with a ratio above 0.85 treated as an acceptable level of traffic-count reproduction.

Parameter-recovery accuracy is evaluated using RMSE and MAPE between the estimated and synthetic ground-truth parameters. These measures are reported separately for the two BPR parameters ($\alpha$ and $\beta$). They should be interpreted as auxiliary diagnostics rather than as the primary calibration criterion. Because the number of unknown link-specific parameters exceeds the number of observed link flows and equilibrium interactions may allow different parameter combinations to produce similar traffic patterns, close reproduction of observed flows does not necessarily imply unique recovery of the generating parameter vector.

Computational efficiency is measured by the total number of complete UE assignment calls required during calibration. UE assignment constitutes the dominant transportation-model computation in the proposed framework, and agents without the differentiable UE layer may require repeated assignments to approximate search directions. Wall-clock time is not used as the primary efficiency metric because it would additionally reflect LLM inference latency, provider-specific implementation, code generation, debugging, and communication overhead that are not intrinsic to the transportation optimization procedure.

Finally, run-to-run consistency is evaluated by independently repeating each agent condition five times using the same task specification, data, and initial conditions. The verified UE solver and adjoint calculation are deterministic for a fixed parameter vector, but the overall agent workflow can still vary because LLM tool use and procedural choices are stochastic. Consistency is therefore assessed using the coefficient of variation (CV) in flow RMSEs across the five runs. A lower CV indicates that the agent more consistently arrives at solutions with comparable calibration accuracy.

Table 4. Evaluation metrics and criteria

| Category | Metric | Remark |
|---|---|---|
| **(1) Flow-reproduction accuracy** | $\mathrm{RMSE} = \sqrt{\frac{1}{\lvert H\rvert}\sum_{a\in H}(x_a^* - x_a^{obs})^2}$ | Reduction relative to the baseline (prior) |
| | $\mathrm{MAPE} = \frac{100}{\lvert H\rvert}\sum_{a\in H}\left\lvert\frac{x_a^* - x_a^{obs}}{x_a^{obs}}\right\rvert$ | Lower is better |
| | $\mathrm{GEH}_a = \sqrt{\frac{2(x_a^* - x_a^{obs})^2}{x_a^* + x_a^{obs}}}$ | Ratio of links with GEH < 5<br>> 0.85 is acceptable |
| **(2) Parameter-recovery accuracy** | $\mathrm{RMSE} = \sqrt{\frac{1}{\lvert H\rvert}\sum_{a\in H}(p_a^* - p_a^{GT})^2}$ | Reduction relative to the baseline (prior) |

| Category | Metric | Remark |
|---|---|---|
| | $\text{MAPE} = \frac{100}{\lvert H\rvert}\sum_{a\in H}\left\lvert\frac{p_a^* - p_a^{GT}}{p_a^{GT}}\right\rvert$ | Lower is better |
| **(3) Computational efficiency** | UE calls | Number of UE assignment calls |
| **(4) Run-to-run consistency** | CV of flow RMSE across five independent runs | Lower is more consistent |

### 4.3 Experimental set-up for Lane-expansion network design under a budget constraint (Task 2)

#### 4.3.1 Definition of Task 2

We consider a lane-expansion network design problem (NDP) formulated on a given directed network $G = (N, L)$. Let $p \in \mathbb{R}^{|L|}$ denote the vector of design variables, where each component $p_a \in [0,1]$ represents the expansion level of link $a \in L$. Here, $p_a = 0$ corresponds to no expansion, while $p_a = 1$ denotes the maximum allowable expansion.

The link capacity under design variable $p$ is given by Eq. (20).

$$C_a(p_a) = C_a^0 + \Delta C_a\, p_a \quad (20)$$

Here, $C_a^0$ is the existing capacity and $\Delta C_a$ is the maximum capacity increase for link $a$.

The BPR function is also chosen as the link performance function and is a function of the design variable of the proposed NDP. Unlike Task 1, the link-specific BPR parameters $\alpha_a, \beta_a > 0$ are held fixed at their predetermined values and are not treated as decision variables.

Eq. (21) denotes the upper-level objective to minimize the total system cost that is the sum of travel time cost and construction cost.

$$\min_p J(p) = \lambda \sum_{a\in L} x_a^*(p)\, t_a(x_a^*(p), p_a) + \sum_{a\in L} c_a p_a \quad (21)$$

Here, $\lambda$ represents the monetary value of travel time.

The budget and bound constraints are denoted by Eq. (22) and Eq. (23), respectively.

$$\sum_{a\in L} c_a\, p_a \le B \quad (22)$$

$$0 \le p_a \le 1, \forall a \in A. \quad (23)$$

Lower-level problem (UE layer) is the UE traffic assignment problem from Beckmann formulation. The equilibrium flow $x^*(p)$ is defined as the solution of the UE problem [Eq. (1)].

Using the proposed framework, differentiable UE layer is represented by a differentiable mapping. The equilibrium flows $x^*(p)$ are the solution of the mapping $\mathcal{F}(y^*(p), p) = 0$. Here, $y = (x, \lambda, \mu)$ satisfies the KKT system of the Beckmann problem.

Eq. (24) denotes the gradient of the upper-level objective and is computed via the adjoint formulation.

$$\nabla_p J = \Phi_p - \psi^\top \mathcal{F}_p \tag{24}$$

Here, $\psi$ is obtained from $\mathcal{F}_y^\top \psi = \Phi_y^\top$.

Differentiable UE layer is represented based on the reduced KKT system. Using the proposed framework, the equilibrium flow is treated as the solution $x^*(p)$ of a differentiable mapping defined implicitly by the reduced KKT system of the Beckmann UE problem.

Let $\mathcal{A}(p) = \{a \in L \mid x_a^*(p) > 0\}$ denote the locally active link set and let $\mathcal{Z}(p) = L \setminus \mathcal{A}(p)$ denote the inactive set. Under the local active-set stability assumption, the complementarity conditions can be eliminated and the KKT system reduces to a smooth system in $(x_{\mathcal{A}}, \lambda)$, $\tilde{\mathcal{F}}(x_{\mathcal{A}}^*(p), \lambda^*(p), p) = 0$, where $\tilde{\mathcal{F}} = \begin{bmatrix} t_{\mathcal{A}}(x^*(p), p) + A_{\mathcal{A}}^\top \lambda^*(p) \\ A_{\mathcal{A}} x_{\mathcal{A}}^*(p) - b \end{bmatrix}$.

The inactive flows are fixed as $x_{\mathcal{Z}}^*(p) = 0$. Therefore, the UE layer is represented by the reduced implicit mapping. Eq. (25) denotes the adjoint gradient.

$$\left(x_{\mathcal{A}}^*(p), \lambda^*(p)\right) = \tilde{\mathcal{F}}^{-1}(0; p), x_{\mathcal{Z}}^*(p) = 0 \tag{25}$$

The adjoint gradient is then computed from Eq. (26) and Eq. (27).

$$\tilde{\mathcal{F}}_y^\top \psi = \Phi_y^\top, \text{ where } y = (x_{\mathcal{A}}, \lambda) \tag{26}$$

$$\nabla_p J = \Phi_p - \psi^\top \tilde{\mathcal{F}}_p, \text{ where, } \tilde{\mathcal{F}}_y = \begin{bmatrix} D_{t,\mathcal{A}} & A_{\mathcal{A}}^\top \\ A_{\mathcal{A}} & 0 \end{bmatrix} \text{ and } D_{t,\mathcal{A}} = \operatorname{diag}\left(\frac{\partial t_a(x_a, p_a)}{\partial x_a}\right)_{a \in \mathcal{A}} \tag{27}$$

This reduced representation avoids explicitly differentiating the complementarity condition and removes the multiplier $\mu$ from the differentiated system.

For the BPR-type function, the derivative with respect to the design variable $p_a$ is denoted by Eq. (28).

$$\frac{\partial t_a(x_a, p_a)}{\partial p_a} = -\, t_a^0\, \alpha\, \beta \left(\frac{x_a}{C_a(p_a)}\right)^\beta \frac{\Delta C_a}{C_a(p_a)} \tag{28}$$

### 4.3.2 Experimental settings for Task 2

Task 2 is evaluated using the same modified Sioux Falls network as Task 1. All 76 directed links are treated as continuous capacity-expansion candidates. For each candidate link, capacity is increased according to its link-specific maximum expansion level, and the corresponding construction cost is incorporated into both the upper-level objective and the capital-budget constraint.

Two demand conditions are considered to examine the performance of the agents under different levels of route interaction. The first, referred to as the "Base demand" scenario, uses the original Sioux Falls OD matrix, with a total demand of approximately 396 thousand vehicle trips. A capital budget of 250 MUSD is imposed. Under this condition, the spatial demand pattern is relatively close to the original benchmark configuration, so highly congested links provide comparatively direct information about potentially beneficial capacity expansions.

The second condition, referred to as the "Complex demand" scenario, is constructed to increase interactions among alternative routes while keeping the total demand and the number of positive-demand OD pairs unchanged. The OD pairs are ranked according to the number of rational candidate paths available between their origins and destinations. The concept of rational path originates from Dial (1971). Seventy-five percent of the total demand is allocated to the upper 25% of OD pairs with relatively more alternative paths, while the remaining 25% is distributed among the other 75% of OD pairs. This redistribution increases the extent to which a capacity change on one link can alter route choices and UE flows elsewhere in the network. A capital budget of 400 MUSD is applied under this condition.

The budget levels in both scenarios are chosen to provide sufficient room for capacity expansion rather than to force the solution to the budget boundary. Consequently, the experiment examines whether an agent can determine not only where additional capacity is beneficial but also when further investment ceases to provide sufficient network-wide benefit.

To maintain heterogeneous congestion responses across links, the link-specific BPR parameters generated for Task 1 are reused in Task 2 and held fixed throughout the network-design experiments. Thus, the design variables in Task 2 are the capacity-expansion levels, whereas the BPR parameters are treated as fixed network attributes.

The four agent configurations defined in Section 4.1 are evaluated again for Task 2. All agents begin from the same initial capacity-expansion design. A run terminates when the relative improvement in the best objective value falls below $10^{-4}$ for three consecutive outer iterations. The same stopping criterion is explicitly provided in the prompt for all agents.

The construction of the search direction depends on the capabilities available to each agent. Proposed-Agent and Gradient-Agent receive the exact adjoint gradient from the differentiable UE layer. UE-Agent has access to the verified UE assignment engine but must construct a search strategy without analytical gradients, whereas Naive-Agent must construct both its transportation-modeling and upper-level search procedures. Step-size selection is left to each agent rather than being prescribed by a common numerical optimizer. The experiment therefore evaluates the complete tool-assisted planning process, including how the LLM uses the information made available to it, rather than comparing only fixed numerical optimization algorithms.

For fair comparison of the final solutions, every submitted capacity-expansion vector is re-evaluated using the same verified UE assignment procedure with a relative-gap tolerance of $5 \times 10^{-7}$. The upper-level objective is then recomputed using this common rigorous UE assignment logic. Accordingly, differences in the reported objective values reflect differences in the submitted network designs rather than differences in the convergence accuracy of the traffic-assignment procedures used internally by the agents.

#### 4.3.3 Evaluation metrics for Task 2

The CNDP experiments are evaluated primarily in terms of solution quality and computing efficiency. Solution quality is measured by the final upper-level objective value obtained after the submitted design is re-evaluated using the common verified UE assignment solver. The objective accounts for both the travel-time cost resulting from the UE traffic pattern and the construction cost associated with the capacity expansions. A lower objective value therefore indicates a more favorable trade-off between congestion reduction and infrastructure investment.

Computing efficiency is measured by the total number of UE assignment calls made during the agent's search before the final design is submitted. A smaller number indicates that the agent reached its submitted solution with fewer UE assignments. This metric is particularly relevant because agents without the differentiable UE layer may require repeated traffic assignments to evaluate perturbations or alternative designs, whereas the differentiable UE layer provides the full objective gradient from the equilibrium solution and the associated adjoint calculation.

The number of UE calls should be interpreted as a measure of transportation-model evaluation effort rather than as a direct measure of wall-clock computational time. The internal implementation of an assignment call may differ across agent configurations, particularly for Naive-Agent, which may construct its own assignment procedure. The metric therefore primarily indicates how intensively each agent relies on repeated UE flow evaluations during the planning process. Taken together, the two metrics characterize both the quality of the final network design and the amount of equilibrium evaluation required to obtain it.

# 5. Results and Discussions

## 5.1 Results of BPR Parameter Calibration (Task 1)

### 5.1.1 Calibration accuracy

Table 5 summarizes the BPR calibration results obtained with Claude Sonnet as the common base LLM. A clear performance separation emerges according to whether the differentiable UE layer is available. Proposed-Agent and Gradient-Agent, both equipped with the differentiable UE layer, reproduce the observed link flows with nearly identical and very small errors. Their performance is substantially better than that of UE-Agent and Naive-Agent, neither of which has access to exact equilibrium gradients.

The similarity between Proposed-Agent and Gradient-Agent is particularly informative. Because the two agents use the same verified UE solver and the same adjoint gradient but differ in whether these capabilities are accessed through MCP or a command-line interface, their nearly identical calibration accuracy indicates that the numerical quality of the final calibration is primarily attributable to the differentiable UE layer rather than to the interface through which it is accessed.

UE-Agent and Naive-Agent perform considerably worse in terms of flow reproduction. Although their relative ranking varies depending on the metric, neither approaches the accuracy of the gradient-equipped agents. The results therefore do not support a general conclusion that either verified UE access alone or an independently constructed assignment procedure is sufficient for reliable high-dimensional calibration. Rather, the dominant distinction is between agents with and without the exact equilibrium sensitivity that the differentiable UE layer provides.

Table 5. BPR parameter calibration performance.

| Metric | | Proposed-Agent | Gradient-Agent | UE-Agent | Naive-Agent |
|---|---|---|---|---|---|
| Flow RMSE | | 7.96 | 7.96 | 198.25 | 155.13 |
| Flow MAPE (%) | | 0.057 | 0.057 | 1.498 | 0.977 |
| GEH $<$ 5 ratio | | 1.000 | 1.000 | 1.000 | 0.983 |
| Parameter RMSE | $\boldsymbol{\alpha}$ RMSE | 0.135 | 0.135 | 0.211 | 0.209 |
| | $\boldsymbol{\beta}$ RMSE | 0.996 | 0.996 | 1.153 | 1.585 |
| Parameter MAPE (%) | $\boldsymbol{\alpha}$ MAPE (%) | 70.2 | 70.2 | 141.3 | 116.4 |
| | $\boldsymbol{\beta}$ MAPE (%) | 20.1 | 20.1 | 23.3 | 31.7 |
| UE assignment calls | | 1,501 | 1,504 | 3,512 | 344,863 |

Fig. 2 reinforces this pattern at the link level. The residuals generated by Proposed-Agent and Gradient-Agent remain tightly concentrated around zero across the monitored links, whereas substantially wider residual distributions are observed for the two gradient-free agents. The advantage of differentiable UE layer is therefore not limited to an aggregate error measure but is reflected across individual traffic-count observations.

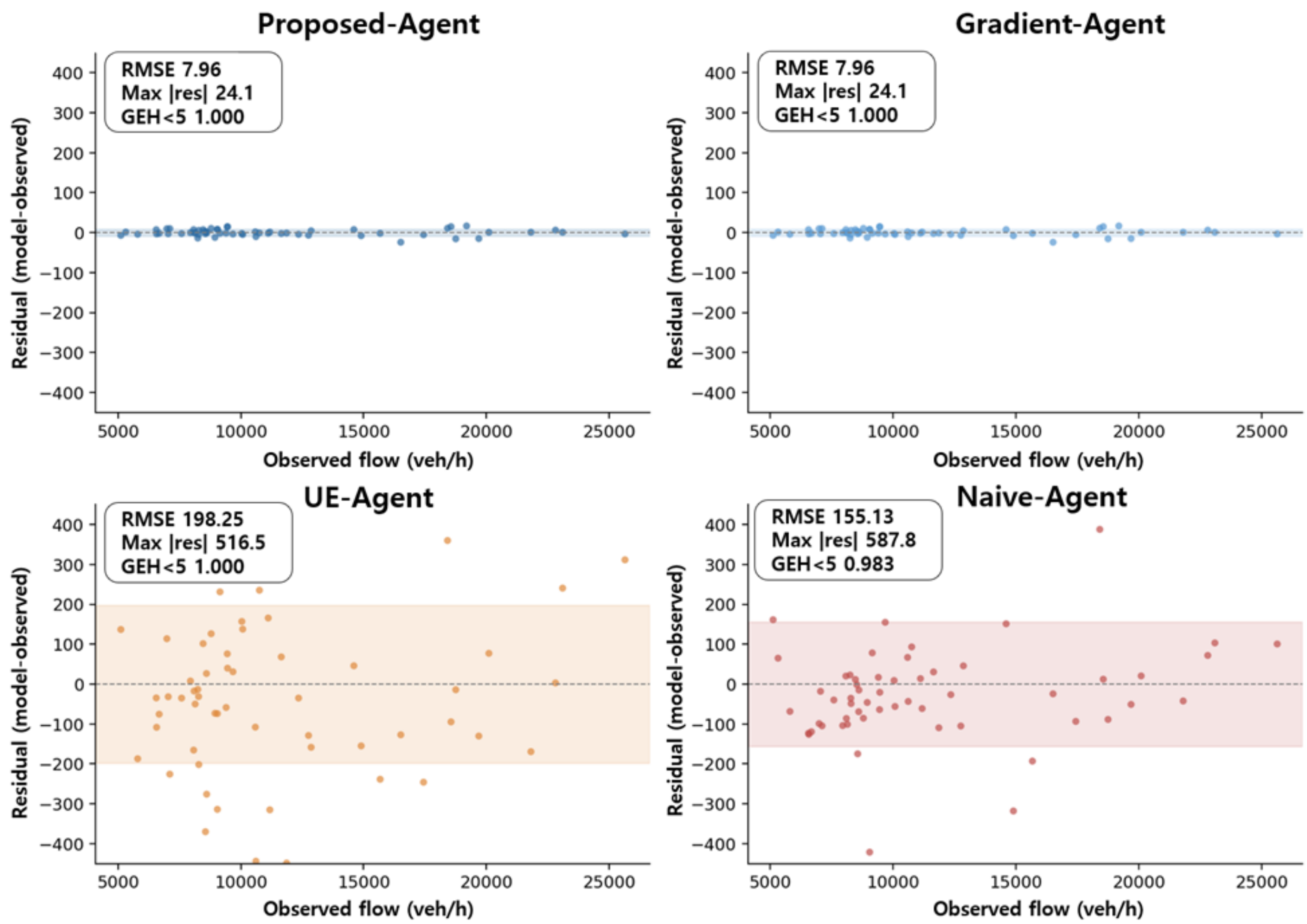

Fig. 2 Reproducibility of link traffic flows

Parameter recovery presents a more nuanced picture. As shown in Table 5 and Fig. 3, the gradient-equipped agents also yield parameter estimates that are generally closer to the synthetic ground truth than those obtained without the differentiable UE layer. However, the separation is considerably less pronounced than for flow reproduction. In particular, substantial dispersion remains in the recovered link-specific BPR coefficients even when the observed flows are reproduced very accurately.

This difference is consistent with the inverse nature of the calibration problem. The optimization objective is defined in terms of observed equilibrium flows rather than direct discrepancies between estimated and ground-truth BPR parameters. Moreover, the number of link-specific parameters exceeds the number of monitored links, so multiple parameter combinations may produce similar UE flow patterns. Consequently, successful reproduction of observed flows does not imply unique recovery of the generating parameter vector. The parameter-recovery measures should therefore be interpreted as auxiliary diagnostics, while flow reproduction remains the primary measure of calibration performance.

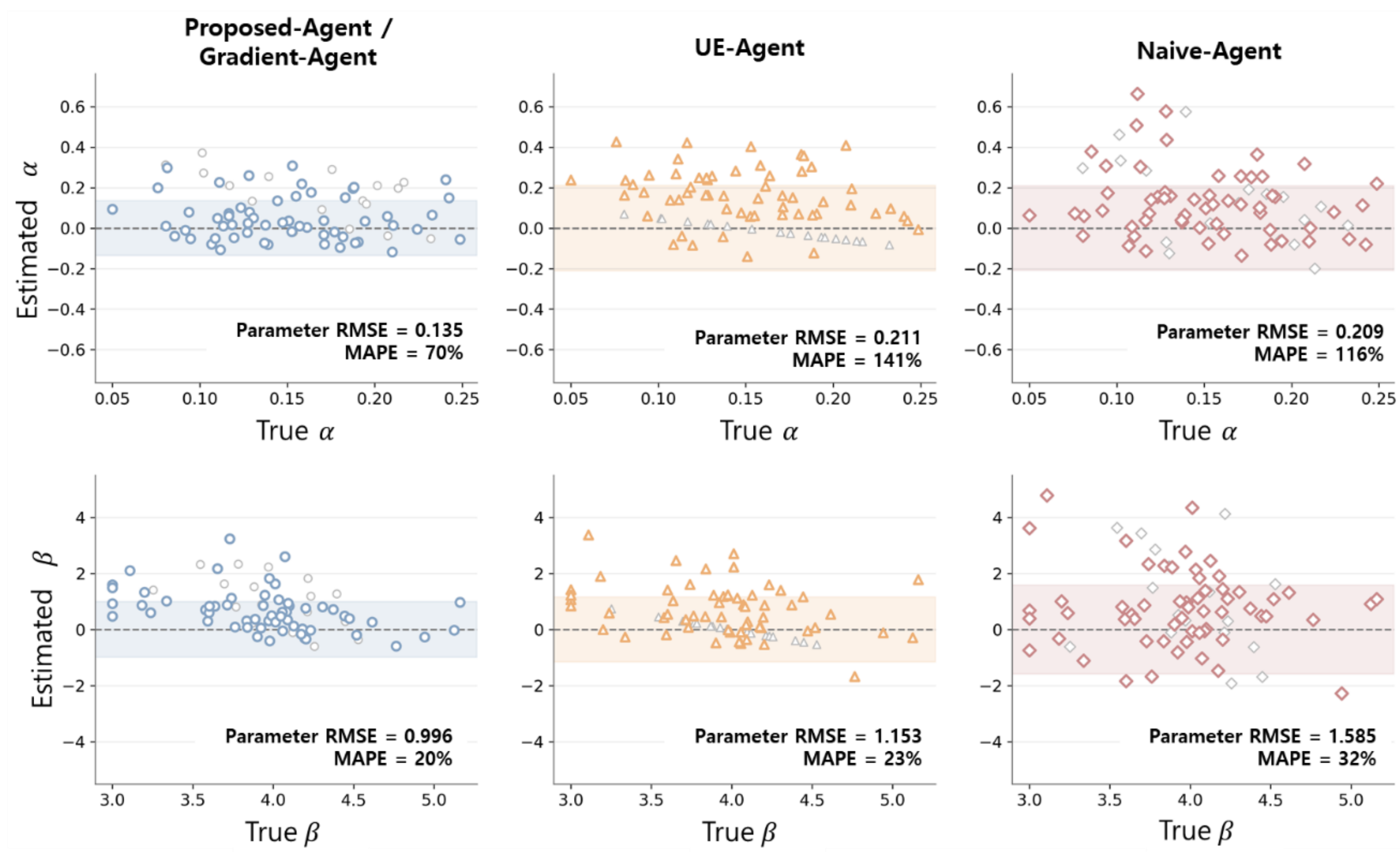

Fig. 3 Reproducibility of BPR parameters

### 5.1.2 Computational efficiency

The computational results in Table 5 further distinguish the gradient-equipped and gradient-free agents. Proposed-Agent and Gradient-Agent require essentially one complete UE evaluation per outer calibration iteration, reflecting the fact that the adjoint system provides the full high-dimensional gradient after the equilibrium solution has been obtained.

In contrast, the two agents without the differentiable UE layer require substantially more UE assignment evaluations. UE-Agent must infer a useful search direction from repeated evaluations of the verified UE assignment model, while Naive-Agent additionally constructs its own assignment and optimization procedures. The resulting increase in UE assignment calls is particularly pronounced for Naive-Agent. Thus, the absence of the differentiable UE assignment layer affects not only calibration accuracy but also the amount of repeated UE flow evaluations required to reach a solution.

An important control result is the near-identical computational effort of Proposed-Agent and Gradient-Agent. For this task, the number of outer iterations is fixed and both agents use the same UE and adjoint routines. Consequently, MCP does not materially reduce the numerical workload of the calibration itself. Its role is instead to provide a structured interface through which the LLM accesses the established transportation demand analysis tools. The computational advantage observed in Task 1 should therefore be attributed primarily to the differentiable UE layer rather than to MCP.

#### 5.1.3 Run-to-Run Consistency

Table 6 and Fig. 4 compare the consistency of the four agent configurations across repeated runs using identical data, initial conditions, and task specifications. The most notable result is the sharp contrast between the gradient-equipped and gradient-free agents.

Proposed-Agent and Gradient-Agent repeatedly reach essentially the same level of flow-reproduction accuracy. In contrast, both UE-Agent and Naive-Agent exhibit substantial variation across independent runs. Thus, the exact gradient from the differentiable UE layer not only improves the average quality of calibration but also substantially reduces the dependence of the final result on the particular search procedure generated during an LLM session.

Table 6. BPR calibration consistency

| Agent | Mean RMSE | CV in RMSEs |
|---|---|---|
| **Proposed-Agent** | 7.96 | 0.00% |
| **Gradient-Agent** | 7.96 | 0.00% |
| **UE-Agent** | 332.82 | 53.7% |
| **Naive-Agent** | 323.63 | 37.0% |

This distinction is important for LLM-assisted transportation planning. The verified UE solver itself is deterministic for a given parameter vector, but an LLM may choose different perturbations, search directions, step sizes, or stopping decisions across sessions. Providing a verified UE solver alone therefore does not eliminate procedural variability at the upper level. The differentiable UE layer supplies a mathematically defined descent direction and thereby constrains a major source of variability in the optimization process.

The relative dispersion of UE-Agent and Naive-Agent should not be interpreted as evidence that one gradient-free strategy is inherently more reliable than the other. Their performance ranges overlap substantially, and both remain markedly less consistent than the gradient-equipped configurations. The stronger conclusion supported by the experiment is therefore that exact UE differentiation stabilizes high-dimensional calibration across repeated LLM executions.

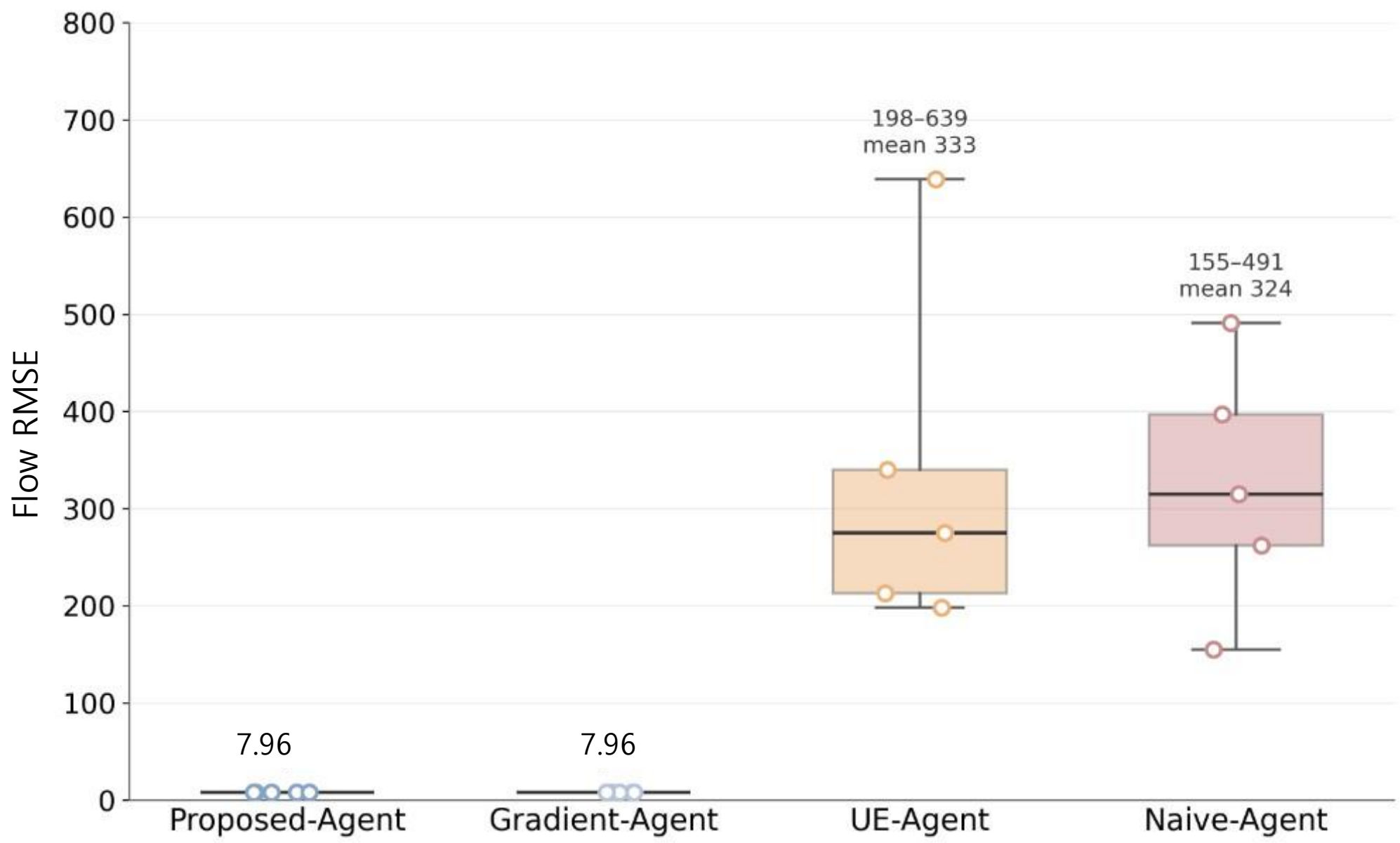

Fig. 4 Flow RMSE variation across five runs

### 5.1.4 Sensitivity to different LLMs

The preceding analyses use Claude Sonnet as the common base model to isolate differences in tool access. Table 7 and Fig. 5 extend the experiment across four LLMs to examine whether the benefit of the differentiable UE layer depends on the capability of the underlying language model.

A strikingly different pattern appears between gradient-equipped and gradient-free agents. Proposed-Agent and Gradient-Agent reach essentially the same verified calibration performance across all four LLMs. In other words, once the differentiable UE layer is supplied as a transportation-domain tool, the final flow-reproduction accuracy becomes largely insensitive to the choice of LLM in this experiment.

The gradient-free agents show the opposite behavior. Their calibration accuracy varies substantially across LLMs, with more capable models generally producing better solutions. This indicates that, without the differentiable UE layer, a larger share of the optimization burden is transferred to the LLM itself: the model must devise numerical perturbations, infer search directions, and organize repeated equilibrium evaluations. Performance consequently depends strongly on the optimization and tool-use capabilities of the underlying language model.

Table 7 Mean flow RMSEs according to different LLMs and agents

| Base LLM model | Proposed-Agent | Gradient-Agent | UE-Agent | Naive-Agent |
|---|---|---|---|---|
| **GPT-Luna** | 7.96 | 7.96 | 677.9 | 631.4 |
| **Claude-Sonnet** | 7.96 | 7.96 | 332.8 | 323.6 |
| **GPT-Sol** | 7.96 | 7.96 | 280.5 | 267.2 |
| **Claude-Opus** | 7.96 | 7.96 | 32.50 | 36.48 |

Increasing LLM capability narrows this gap but does not eliminate it. Even the strongest gradient-free configurations tested here remain less accurate than the agents supplied with the differentiable UE layer. The results therefore suggest that improvements in general-purpose LLM capability and the provision of mathematically grounded domain tools are not substitutes. A stronger LLM can improve an improvised optimization procedure, whereas a differentiable equilibrium tool directly supplies the transportation-specific numerical information required for the task.

More broadly, the cross-LLM experiment supports the architectural premise of the proposed framework. The LLM is most effectively used to interpret the planning task and orchestrate the computational workflow, while equilibrium solution and sensitivity analysis are delegated to verified transportation models. Such separation makes the resulting planning performance less dependent on the numerical optimization capability of a particular LLM and more reproducible across different LLM implementations.

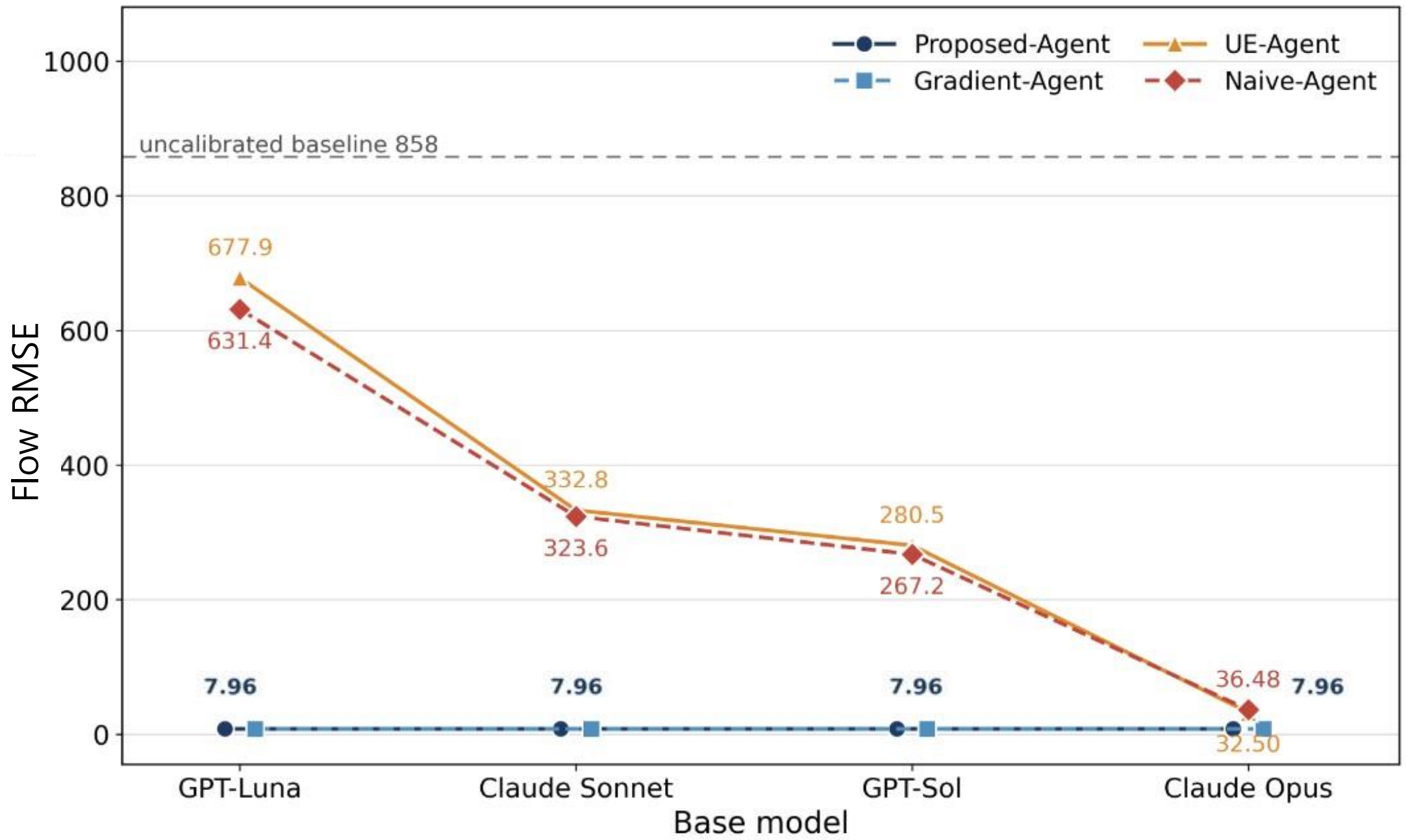

Fig. 5 Flow RMSEs according to different LLM models

### 5.2 Results of the Continuous Network Design Problem (Task 2)

#### 5.2.1 Base Demand Scenario

Table 8 and Fig. 6 compare the four agent configurations across the four different LLMs under the Base demand scenario. In contrast to the BPR calibration task, the final network-design objectives are relatively close across most agent configurations. This indicates that, under the original Sioux Falls demand pattern, much of the useful capacity-expansion structure can be identified even without exact differentiation of UE assignment flows. Highly congested links provide relatively informative signals regarding where additional capacity may be beneficial, allowing gradient-free agents to construct reasonably competitive designs.

Nevertheless, a clearer separation appears in the amount of UE evaluation required to obtain those designs. Across all reference LLMs, Proposed-Agent reaches its submitted solution with substantially fewer UE assignment calls than the other configurations. Gradient-Agent generally requires more evaluations than Proposed-Agent but fewer than the agents without the differentiable UE layer. Thus, under the Base scenario, the main contribution of the differentiable UE layer is observed less in the final objective value than in the efficiency with which the LLM navigates the design space.

The comparison between Proposed-Agent and Gradient-Agent provides an additional insight. Because both agents receive the same verified UE solution and exact adjoint gradient, their similar final objective values are expected. Their difference lies primarily in how these numerical capabilities are exposed to and orchestrated by the LLM. The consistently smaller number of equilibrium evaluations used by Proposed-Agent suggests that the structured MCP workflow can reduce redundant or unsuccessful model evaluations during adaptive search. This result should not be interpreted as MCP reducing the computational cost of an individual UE or adjoint calculation; rather, it indicates greater efficiency in coordinating repeated transportation-model evaluations.

The relatively small differences in final objective value also suggest that the Base scenario does not strongly discriminate between alternative search mechanisms. In this setting, approximate congestion information may already reveal many of the attractive expansion candidates. Exact UE differentiation becomes more important when the marginal network-wide benefit of a capacity change is difficult to infer directly from the current congestion pattern, as examined in the following complex scenario.

Table 8 Performance of agents varying LLMs under "Base Scenario"

| LLMs | Objective function value (MUSD) | | | | Number of UE assignment calls | | | |
|---|---|---|---|---|---|---|---|---|
| | Proposed-Agent | Gradient-Agent | UE-Agent | Naive-Agent | Proposed-Agent | Gradient-Agent | UE-Agent | Naive-Agent |
| GPT-Luna | 371.4 | 372.1 | 372.1 | 376.4 | 93 | 519 | 946 | 1,488 |
| Claude-Sonnet | 371.7 | 372.1 | 373.8 | 372.7 | 73 | 105 | 279 | 1,082 |
| GPT-Sol | 371.1 | 371.6 | 372.0 | 371.8 | 80 | 145 | 505 | 1,762 |
| Claude-Opus | **371.0** | 371.6 | 374.5 | 371.6 | **73** | 132 | 263 | 1,436 |

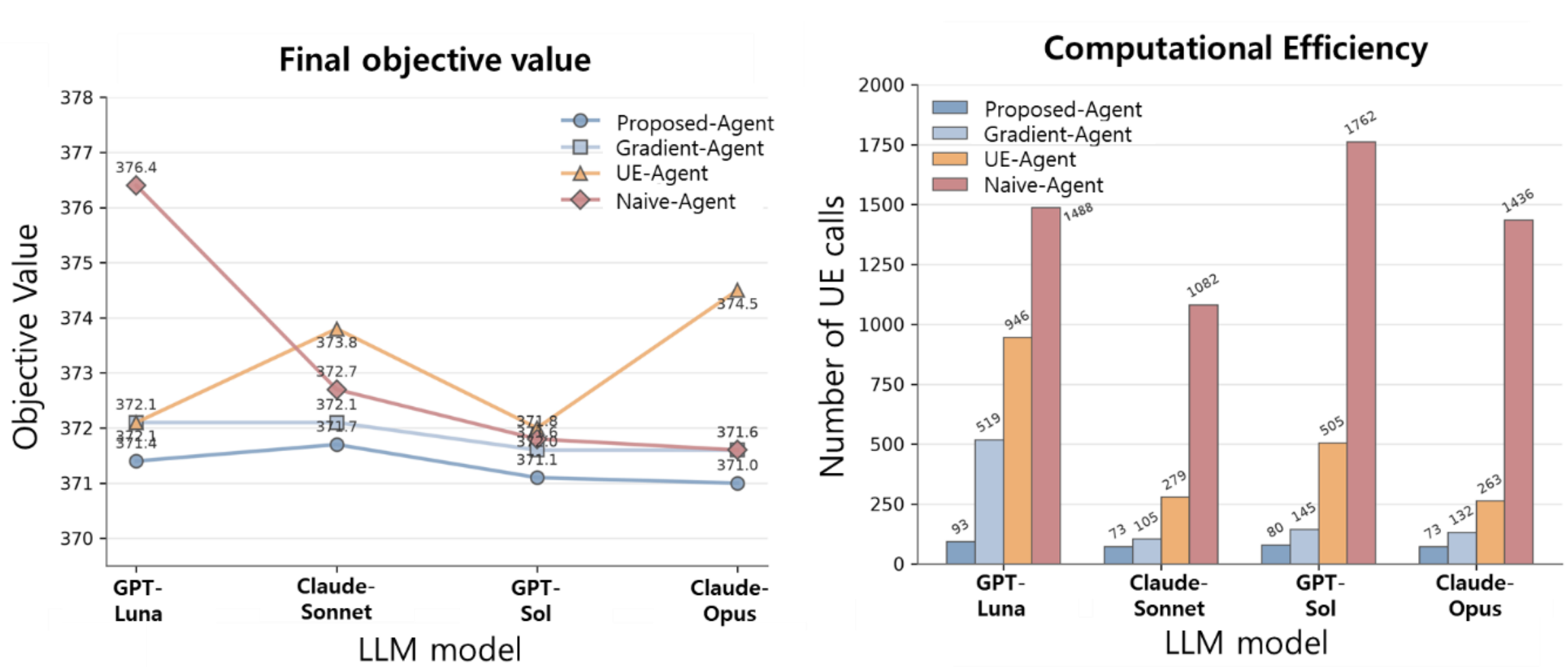

Fig. 6 Visual presentation of the performance of agents by varying LLMs under "Base scenario"

### 5.2.2 Solutions for "Complex demand" scenario

Table 9 and Fig. 7 report the results under the Complex demand scenario, in which demand is concentrated on OD pairs with a larger number of alternative paths. Compared with the Base scenario, a more pronounced separation in solution quality emerges between agents with and without the differentiable UE layer, although the magnitude of the difference varies across reference LLMs.

Proposed-Agent and Gradient-Agent consistently remain among the best-performing configurations. In contrast, the gradient-free agents become more sensitive to the capability of the underlying LLM and, for some models, produce substantially poorer network designs. The deterioration is particularly pronounced for weaker LLM models, whereas a stronger LLM can partially compensate for the absence of exact gradient information. Even so, the gradient-equipped configurations remain more stable across LLMs.

This pattern is consistent with the increased route interaction introduced in the Complex demand scenario. When multiple alternative routes compete for the same OD demand, expanding one link may redistribute equilibrium flows across several other links. Consequently, the benefit of an expansion cannot be inferred reliably from the congestion level of that link alone. Exact equilibrium sensitivity provides information on these network-wide responses and therefore becomes more useful for distinguishing capacity changes that generate genuine system-level improvements from those whose apparent local benefits are offset by equilibrium redistribution.

The comparison between Proposed-Agent and Gradient-Agent again shows little systematic difference in final solution quality measured by objective function values, which is consistent with their access to the same exact gradient. However, Proposed-Agent requires fewer UE evaluations across all four LLMs. As in the Base scenario, this difference is best interpreted as an effect on workflow efficiency rather than on the numerical quality of the gradient itself. The MCP interface provides the LLM with structured transportation operations, state information, and stopping logic, thereby reducing the amount of repeated interaction with the underlying equilibrium model required to complete the search.

Taken together, the two demand scenarios reveal a complementary role of the proposed components. Under the relatively straightforward Base condition, the differentiable UE layer primarily reduces the effort required to obtain a competitive design, while improvements in final objective value remain modest. As route interactions become stronger, exact equilibrium sensitivity contributes more visibly to solution quality as well. MCP, in turn, does not alter the UE gradient but appears to improve the efficiency with which the LLM uses that information during an adaptive network-design workflow.

These findings suggest that the value of domain-specific differentiable tools depends on the structure of the planning problem. When useful interventions can be inferred largely from observable congestion patterns, a capable LLM may approximate the relevant design decisions without explicit differentiation information. When the effects of an intervention propagate through multiple competing routes, however, local heuristics become less informative and mathematically grounded UE gradient provides a more reliable basis for network-level decision making.

Table 9 Performance of agents varying LLMs under “Complex scenario”

| LLMs | Objective function value (MUSD) | | | | Number of UE assignment calls | | | |
|---|---|---|---|---|---|---|---|---|
| | Proposed-Agent | Gradient-Agent | UE-Agent | Naive-Agent | Proposed-Agent | Gradient-Agent | UE-Agent | Naive-Agent |
| GPT-Luna | 622.9 | 622.9 | 716.9 | 1,873.9 | 80 | 119 | 122 | 155 |
| Claude-Sonnet | 622.0 | 622.0 | 622.1 | 622.3 | 84 | 128 | 180 | 1,487 |
| GPT-Sol | 622.5 | 622.6 | 624.4 | 696.8 | 97 | 180 | 209 | 404 |
| Claude-Opus | **620.5** | 620.7 | 621.9 | 625.3 | **82** | 92 | 149 | 160 |

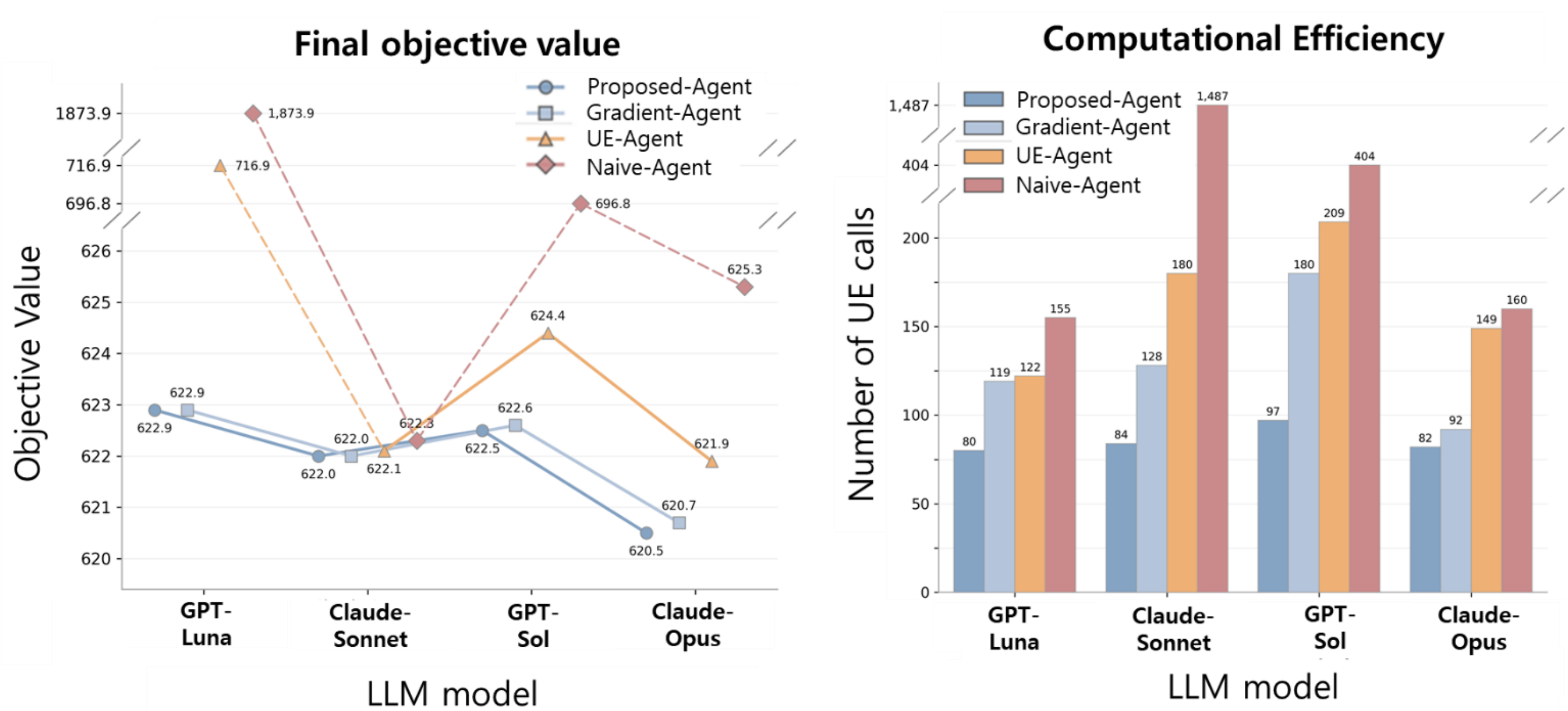

Fig. 7 Visual presentation of the performance of agents by varying LLMs under “Complex scenario”

## 6. Conclusions

This study developed an LLM-assisted transportation planning framework that integrates natural-language interaction, MCP-based tool orchestration, a verified traffic-assignment engine, and a differentiable UE layer. The framework is designed so that the LLM interprets planning objectives and coordinates the workflow, while traffic assignment, equilibrium sensitivity, and feasibility checks are delegated to mathematically grounded transportation models.

The differentiable UE layer is constructed by directly differentiating the reduced KKT system of the Beckmann formulation under local active-set stability. This provides equilibrium sensitivities without differentiating through the iterations of a particular assignment algorithm, allowing the gradient computation to remain independent of the forward UE solver.

The two experiments demonstrate complementary benefits of this architecture. In BPR parameter calibration, agents equipped with the differentiable UE layer achieve substantially better flow reproduction, require fewer repeated equilibrium evaluations, and exhibit much higher run-to-run consistency than gradient-free agents. Their performance is also nearly invariant across the tested LLMs, whereas gradient-free agents depend strongly on LLM capability. This indicates that providing the exact transportation-domain gradient substantially reduces the numerical optimization burden placed on the LLM.

The network-design experiment shows a different pattern. Under the Base demand scenario, most agents obtain similar design quality, while the differentiable UE layer mainly improves search efficiency. Under the more route-interactive Complex demand scenario, exact equilibrium sensitivity becomes more important for solution quality because capacity changes induce broader equilibrium-flow redistribution that cannot be inferred reliably from local congestion alone.

The experiments also clarify the role of MCP. MCP does not improve the mathematical UE solution or gradient itself. Instead, its contribution is primarily at the workflow level. In the adaptive CNDP task, MCP-based access reduces the number of repeated equilibrium evaluations by providing structured tool calls, feasibility checks, state information, and stopping logic to the LLM.

Overall, the results suggest that LLMs are most effective when used as planners and orchestrators rather than as substitutes for transportation optimization models. A stronger LLM can partially compensate for missing numerical tools, but verified equilibrium and its gradient modules provide substantially more reliable and consistent performance.

Several limitations remain. The experiments use the Sioux Falls benchmark, the calibration data are synthetic, the UE gradient is local to a fixed active set, and the continuous network-design problem does not guarantee a global optimum. Future work should therefore examine larger real-world networks and observed data, discrete design decisions, and broader MCP tool environments.

## Acknowledgement

This research was supported by the Chung-Ang University Research Scholarship Grants in 2025 and, in part, by a Korea Agency for Infrastructure Technology Advancement (KAIA) grant funded by the Ministry of Land, Infrastructure and Transport (grant no. RS-2026-25530752) and also the National Research Foundation of Korea (NRF) Grant funded by the Korean Government (RS-2024-00337956).